\documentclass[11pt]{article}

\usepackage[margin=1in]{geometry}
\usepackage{booktabs}
\usepackage{graphicx}
\usepackage{hyperref}
\usepackage{longtable}
\usepackage{url}
\usepackage{placeins}
\usepackage{tikz}
\usetikzlibrary{positioning, arrows.meta, shapes.geometric, calc, fit, backgrounds}
\definecolor{openweight}{HTML}{4477AA}

\graphicspath{{figures/}}

\title{FormationEval, an open multiple-choice benchmark for petroleum geoscience}
\author{
Almaz Ermilov\\
UiT The Arctic University of Norway\\
\texttt{almaz.ermilov@gmail.com}
}
\date{Preprint}

\begin{document}
\maketitle

\begin{abstract}
This paper presents FormationEval, an open multiple-choice question benchmark for evaluating language models on petroleum geoscience and subsurface disciplines. The dataset contains 505 questions across seven domains including petrophysics, petroleum geology and reservoir engineering, derived from three authoritative sources using a reasoning model with detailed instructions and a concept-based approach that avoids verbatim copying of copyrighted text. Each question includes source metadata to support traceability and audit. The evaluation covers 72 models from major providers including OpenAI, Anthropic, Google, Meta and open-weight alternatives. The top performers achieve over 97\% accuracy, with Gemini 3 Pro Preview reaching 99.8\%, while tier and domain gaps persist. Among open-weight models, GLM-4.7 leads at 98.6\%, with several DeepSeek, Llama, Qwen and Mistral models also exceeding 93\%. The performance gap between open-weight and closed models is narrower than expected, with several lower-cost open-weight models exceeding 90\% accuracy. Petrophysics emerges as the most challenging domain across all models, while smaller models show wider performance variance. Residual length bias in the dataset (correct answers tend to be longer) is documented along with bias mitigation strategies applied during construction. The benchmark, evaluation code and results are publicly available.
\end{abstract}

\section{Introduction}

Large language models (LLMs) are increasingly applied to domain-specific tasks in science and engineering, yet their capabilities in specialized fields remain difficult to assess. General benchmarks like MMLU \cite{mmlu} cover broad knowledge but offer limited focus on specialized fields. For petroleum geoscience and subsurface engineering (fields requiring understanding of well logging physics, reservoir characterization and geological interpretation), publicly available benchmarks remain limited.

This work addresses the gap with FormationEval, a 505-question multiple-choice benchmark covering seven domains: petrophysics, petroleum geology, geophysics, reservoir engineering, sedimentology, drilling engineering and production engineering. Questions are derived from authoritative textbooks and open courseware using a concept-based methodology that tests understanding rather than phrase recognition, while respecting copyright constraints.

The contributions are: (1) a methodology for generating multiple-choice questions (MCQs) from technical sources without verbatim copying; (2) a curated dataset with source metadata and contamination risk labels; and (3) an evaluation of 72 language models across multiple providers, revealing performance patterns by domain and difficulty level.

\section{Related work}

General-purpose benchmarks like MMLU \cite{mmlu} evaluate broad knowledge across 57 subjects but provide limited coverage of specialized domains. MMLU-Pro \cite{mmlu_pro} addresses some limitations with harder reasoning questions yet remains domain-general. For science, ARC \cite{arc} covers grade-school science questions and SciBench \cite{scibench} targets college-level problem solving in physics, chemistry and mathematics. GPQA \cite{gpqa} provides 448 graduate-level questions where even domain experts achieve only 65\% accuracy. Domain-specific benchmarks exist for medicine (MedQA \cite{medqa}, 12,723 questions from medical licensing exams) and law (LegalBench \cite{legalbench}). However, no public MCQ benchmark exists specifically for petroleum geoscience or subsurface engineering disciplines.

Recent work on large language models for geoscience has produced domain-adapted models such as K2 \cite{k2_geoscience}, which further pre-trains LLaMA on 5.5 billion tokens of geoscience text and introduces GeoBench for evaluation, and GeoGalactica \cite{geogalactica}, a 30-billion parameter model trained on 65 billion tokens of geoscience literature. Hadid et al.\ \cite{hadid_genai_geoscience} survey generative AI applications across geoscience disciplines. These efforts demonstrate growing interest in AI for earth sciences but focus on model development rather than standardized evaluation of existing models across petroleum-specific knowledge areas.

Automatic MCQ generation from text is a well-studied problem. Ch and Saha \cite{mcq_survey} survey methods spanning sentence selection, key extraction, question formation and distractor generation. Recent LLM-based approaches include MCQG-SRefine \cite{mcqg_srefine}, which uses iterative self-critique to generate medical exam questions, and various distractor generation techniques surveyed by Alhazmi et al.\ \cite{distractor_survey}. Most existing methods transform source sentences into questions through syntactic manipulation or paraphrasing. The concept-based methodology presented here differs by generating questions from extracted concepts rather than from specific sentences, which avoids close paraphrasing of copyrighted text while testing understanding rather than phrase recognition.

\section{Benchmark design and construction}

\subsection{Task definition and scope}

FormationEval uses a four-choice multiple-choice question format with exactly one correct answer per question. This format is compatible with standard evaluation frameworks and enables straightforward accuracy computation.

Questions cover seven domains: Petrophysics (well logging, formation evaluation), Petroleum Geology (source rocks, migration, trapping), Sedimentology (depositional environments, diagenesis), Geophysics (seismic interpretation, rock physics), Reservoir Engineering (fluid flow, recovery mechanisms), Drilling Engineering (wellbore stability, operations) and Production Engineering (completions, artificial lift). Questions can belong to more than one domain.

Difficulty levels reflect educational background. Easy questions (undergraduate) test definitions and direct recall. Medium questions (graduate or professional) require applying concepts to scenarios. Hard questions (specialist) involve integrating multiple concepts or edge cases.


\subsection{Source selection and licensing policy}

The benchmark draws from three sources: Ellis \& Singer's \textit{Well Logging for Earth Scientists} \cite{ellis_singer_2007} (219 questions), Bj{\o}rlykke's \textit{Petroleum Geoscience} \cite{bjorlykke_2010} (262 questions) and TU Delft OpenCourseWare \cite{tudelft_ocw_2008} (24 questions).

The benchmark uses a concept-based derivation approach. Questions are written from scratch based on concepts extracted from source material, without copying sentences or closely paraphrasing distinctive problem structures. This respects the legal distinction between ideas (not copyrightable) and expression (protected). Standard technical terms (porosity, Archie equation, neutron-density crossplot) may appear as-is since terminology is not copyrightable.

All generated items are tagged with \texttt{derivation\_mode: concept\_based} and include source tracking fields that enable verification without reproducing protected text.


\subsection{Schema and metadata}

Each question includes required fields: unique identifier, question text, four choices, answer index (0--3), answer key (A--D), difficulty level, domains, topics and a rationale explaining the correct answer. The rationale serves dual purposes. It aids human verification during development and provides educational value to benchmark users.

Each item also includes a \texttt{contamination\_risk} label indicating likelihood that similar questions exist in LLM training data: \textit{low} for novel questions specific to the source, \textit{medium} for common concepts where similar questions may exist and \textit{high} for standard introductory topics almost certainly present in training data. Assessing contamination risk is hard without access to training data; this label is an estimate based on topic commonality and cannot be directly checked. Still, approximate labels help when comparing results by risk level.

Provenance metadata includes source identifier, title, chapter reference, license and retrieval date. See Appendix A for the complete schema reference.


\subsection{Multiple-choice question generation pipeline}

Questions are generated using a reasoning model (GPT-5.2 \cite{gpt52} with extra high reasoning effort) that processes source chapters (Section 3.2) through extended chain-of-thought before producing output. The pipeline consists of four stages:

\begin{enumerate}
\item \textbf{Text extraction} converts source PDFs to Markdown using optical character recognition (OCR), preserving structure and mathematical notation.
\item \textbf{Chunking} splits documents by chapter or section, with each chunk sized for model context (typically one chapter, approximately 10,000--15,000 tokens).
\item \textbf{Candidate generation} uses GPT-5.2 with extra high reasoning effort, which receives the chapter text along with a detailed system prompt specifying schema requirements, concept-based derivation rules, difficulty targets and output format. The model generates 5--12 questions per chapter.
\item \textbf{Verification} checks schema compliance (no duplicate choices, answer index in range) and confirms support in source text, flagging ambiguous items.
\end{enumerate}

\begin{figure}[htbp]
  \centering
  \includegraphics[width=\textwidth]{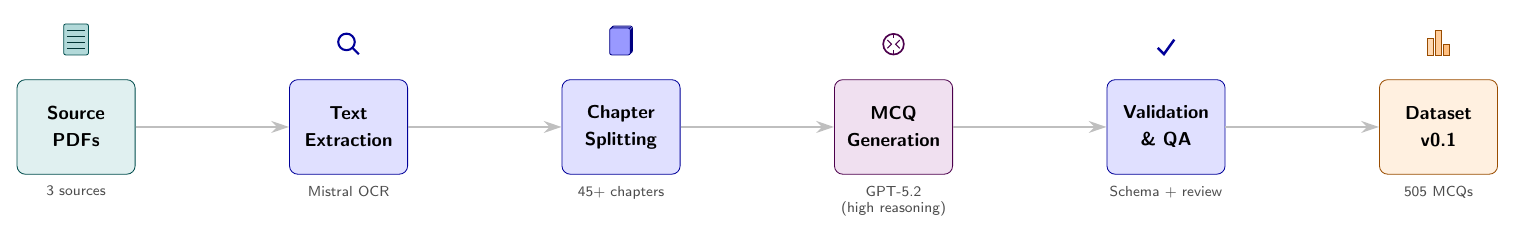}
  \caption{MCQ generation pipeline. Source PDFs are converted to Markdown via OCR,
           split into chapter chunks, processed by GPT-5.2 to generate candidate questions
           and verified for schema compliance and source evidence.}
  \label{fig:mcq_pipeline}
\end{figure}

Figure~\ref{fig:mcq_pipeline} illustrates this pipeline from source ingestion to final dataset.

The system prompt emphasizes that questions must be standalone, answerable from domain knowledge without access to the source chapter. Phrases like ``according to the chapter'' or ``the text describes'' are explicitly prohibited. A summary of the generation prompt is provided in Appendix~B; the full system prompt is available in the repository at \href{https://github.com/AlmazErmilov/FormationEval-an-Open-Benchmark-for-Oil-Gas-Geoscience-MCQ-Evaluation/blob/v0.1/src/prompts/mcq_generator_system_prompt.txt}{\texttt{src/\allowbreak{}prompts/\allowbreak{}mcq\_generator\_system\_prompt.txt}}.


\subsection{Quality assurance and audit}

Human spot-checking is essential despite high LLM instruction-following reliability. All manual verification was performed by the author, a petrophysicist with background in well logging, formation evaluation and petroleum geology. For each source chapter, 2--5 questions were verified against the source material to confirm: (1) the marked answer is correct and unambiguous; (2) distractors are plausible but clearly wrong; (3) no copied phrases appear in questions or answers; (4) the rationale supports the marked answer. This domain expertise was particularly important for petrophysics questions (54\% of the dataset), where assessing answer correctness requires familiarity with logging tool physics and interpretation methods.

The full dataset (505 questions across 45+ chapters) was reviewed in batches grouped by source and chapter, with each batch cross-checked against the source material. The final audit flagged one issue out of 505 questions (a question where the source text contained internally inconsistent units). All other batches passed with no issues. In addition to human spot-checks, LLM-based batch review flagged potential inconsistencies between rationales and correct answers.

Post-generation corrections addressed two categories of issues. First, 55 questions required rewriting to remove chapter self-references (phrases like ``according to the chapter'' or ``the text describes'') that violated the standalone requirement. These were replaced with domain context so that each question is answerable from general petroleum geoscience knowledge. Second, 12 grammar corrections were applied after batch bias-mitigation edits introduced broken sentences.

The generation prompt evolved through eight major iterations over the course of the project. The list of explicit prohibitions (covering self-references, negative phrasing, length balance and qualifier word patterns) grew from 5 to 15 rules. Few-shot examples were also corrected after analysis revealed that two of three examples had the longest option as the correct answer, contradicting the length balance guidance in the prompt text.

The required rationale field accelerated verification throughout this process. When the model produced a plausible-sounding answer but could not write a coherent rationale, the answer was often incorrect or fabricated. This made the rationale field an effective early indicator of generation errors during both automated and manual review.


\subsection{Bias analysis and mitigation}

Initial analysis revealed two exploitable patterns.

\textbf{Length bias} was significant. Correct answers were uniquely longest in over 55\% of questions (expected 25\%), averaging 86.6 characters versus 69.8 for distractors.

\textbf{Qualifier word bias} was also present. Absolute words like ``always'' appeared only in distractors (49 instances, 0\% correct rate), while hedged words like ``may'' appeared only in correct answers (13 instances, 100\% correct rate).

Combined, these patterns could be exploited well above random chance.

\textbf{Mitigation} involved expanding 136 distractors with technical context to reduce length imbalance (from over 55\% to 43.2\% uniquely-longest-is-correct). All ``always'' instances were replaced with varied synonyms (invariably, necessarily, inherently) to break the single-word exploit. The word ``may'' was added to 13 distractors (e.g., ``has no effect'' $\to$ ``may have no effect'') to balance hedging language.

\textbf{Residual issues} remain. Length bias is still above the 25\% baseline. The absolute-word synonyms all have 0\% correct rate, making a combined ``any-absolute-word=wrong'' heuristic still partially exploitable.

For benchmarking purposes, these residual biases should not significantly affect relative comparisons. Since the same patterns apply to all questions, any bias-based advantage would affect all models equally, preserving the validity of model-to-model comparisons. The major exploits (length as a proxy for correctness, ``always'' as a marker for wrong answers, ``may'' as a marker for correct answers) have been addressed. After mitigation, correct answers average 87 characters versus 74 for distractors, a difference of roughly 13 characters (2--3 words). With the uniquely longest answer correct only 43.2\% of the time, a length-based strategy would fail more often than it succeeds. The remaining issues with absolute-word synonyms affect fewer than 10\% of questions. These limitations are documented for transparency; see Appendix~C for detailed before/after metrics.


\subsection{PDF export for review}

For easier review, the dataset is exported to PDF with a cover page, question cards showing all metadata and bookmarks for navigation. This format is more accessible than raw JSON for domain experts performing quality checks and enables browsing questions by domain or topic without programming tools.
An accompanying PDF rendering, generated from the JSON release, provides a readable format for browsing and spot checks \cite{formationeval_dataset,formationeval_pdf}.


\subsection{Community feedback through interactive quiz}

Since all manual verification was performed by a single domain expert (spot checks of 2--5 questions per source chapter, Section~3.5), scalable quality assurance requires additional feedback channels. To address this, an interactive website was developed alongside the benchmark \cite{formationeval_website}. The website includes a quiz mode where domain experts answer benchmark questions and compare their scores against the 72 evaluated language models. This gamified format is designed to encourage engagement: after completing a quiz, users see which models they outperformed or matched, with options to share results on social media. The goal is to attract domain experts who might not otherwise review a static dataset.

Each question in both the quiz results and the question browser includes a feedback button that opens a pre-filled email with the question identifier attached. This allows experts to report incorrect answers, ambiguous phrasing or other issues directly while reviewing a specific question. The question browser provides access to all 505 questions with filtering by domain and difficulty, answer rationales and source citations, enabling systematic review without programming tools.

This approach treats community engagement as a continuous quality assurance process. As more domain experts interact with the benchmark through the quiz, the likelihood of identifying remaining errors increases. Feedback collected through this channel will inform corrections in future versions of the dataset.

The website is available at \url{https://www.formationeval.no}.

\section{Dataset summary}

\begin{figure}[htbp]
  \centering
  \includegraphics[width=\textwidth]{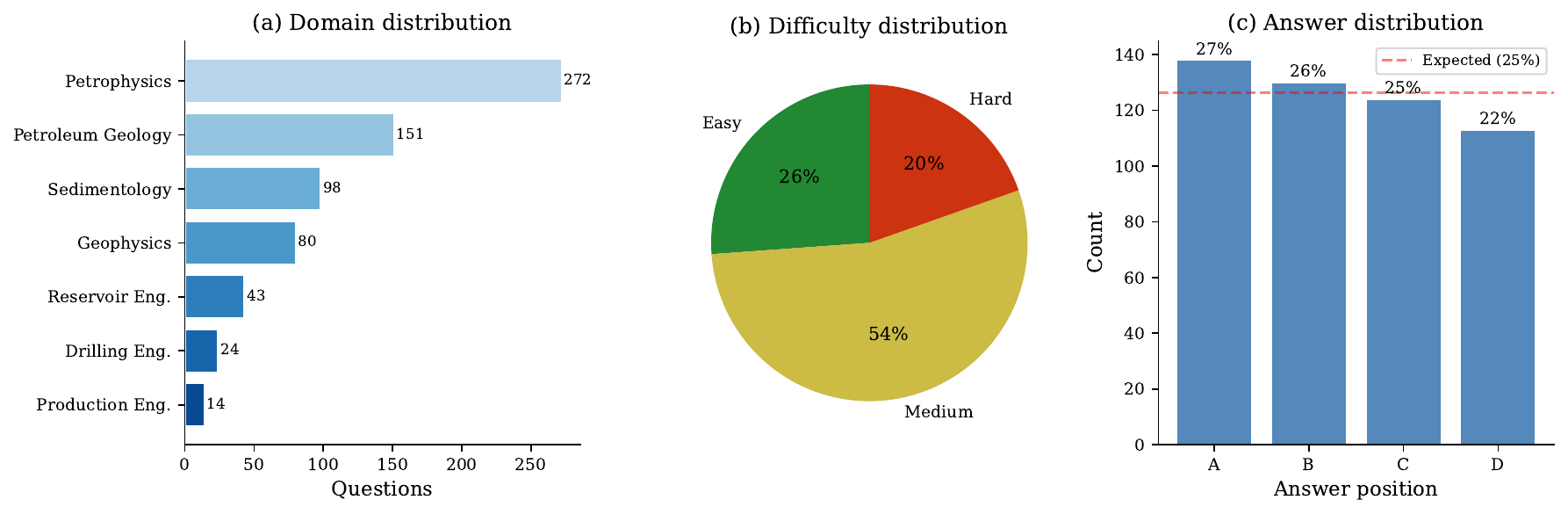}
  \caption{Dataset composition. (a) Questions by domain, with Petrophysics dominating due to source coverage; (b) difficulty distribution close to 30/50/20 targets; (c) answer position distribution near the expected 25\% baseline.}
  \label{fig:dataset_composition}
\end{figure}

Version 0.1 of FormationEval contains 505 questions in English covering 811 unique topics (as questions can address multiple topics). Table~\ref{tab:summary} summarizes key metrics, Table~\ref{tab:distribution} shows the distribution by domain and difficulty and Table~\ref{tab:diff_domain} provides the difficulty breakdown within each domain. Questions are distributed across three sources: Bj{\o}rlykke's textbook contributes the largest share (262 questions, 52\%), followed by Ellis \& Singer (219 questions, 43\%) and TU Delft open courseware (24 questions, 5\%). Figure~\ref{fig:dataset_composition} visualizes these distributions.

Petrophysics represents the largest domain (54\% of questions) reflecting the depth of coverage in the well logging textbook. The difficulty distribution targets 30\% easy, 50\% medium and 20\% hard; the actual distribution (26\%/54\%/20\%) is close to these targets. Answer positions are balanced: A=27\%, B=26\%, C=25\%, D=22\%.

\begin{table}[htbp]
  \centering
  \caption{Dataset summary (v0.1).}
  \label{tab:summary}
  \begin{tabular}{ll}
    \toprule
    Metric & Value \\
    \midrule
    Questions & 505 \\
    Sources & 3 \\
    Domains & 7 \\
    Unique topics & 811 \\
    Language & English \\
    \bottomrule
  \end{tabular}
\end{table}

\begin{table}[htbp]
  \centering
  \caption{Domain and difficulty distribution. Domain counts are non-exclusive (questions may belong to multiple domains).}
  \label{tab:distribution}
  \begin{tabular}{lrr}
    \toprule
    Category & Count & Share \\
    \midrule
    \multicolumn{3}{l}{\textit{By domain}} \\
    Petrophysics & 272 & 54\% \\
    Petroleum Geology & 151 & 30\% \\
    Sedimentology & 98 & 19\% \\
    Geophysics & 80 & 16\% \\
    Reservoir Engineering & 43 & 9\% \\
    Drilling Engineering & 24 & 5\% \\
    Production Engineering & 14 & 3\% \\
    \midrule
    \multicolumn{3}{l}{\textit{By difficulty}} \\
    Easy & 132 & 26\% \\
    Medium & 274 & 54\% \\
    Hard & 99 & 20\% \\
    \bottomrule
  \end{tabular}
\end{table}

\begin{table}[htbp]
  \centering
  \caption{Difficulty distribution by domain. Domain counts are non-exclusive (questions may belong to multiple domains). Percentages show within-domain distribution.}
  \label{tab:diff_domain}
  \begin{tabular}{lrrrr}
    \toprule
    Domain & Easy & Medium & Hard & Total \\
    \midrule
    Petrophysics & 49 (18\%) & 159 (58\%) & 64 (24\%) & 272 \\
    Petroleum Geology & 50 (33\%) & 70 (46\%) & 31 (21\%) & 151 \\
    Sedimentology & 30 (31\%) & 54 (55\%) & 14 (14\%) & 98 \\
    Geophysics & 19 (24\%) & 44 (55\%) & 17 (21\%) & 80 \\
    Reservoir Engineering & 13 (30\%) & 25 (58\%) & 5 (12\%) & 43 \\
    Drilling Engineering & 10 (42\%) & 11 (46\%) & 3 (12\%) & 24 \\
    Production Engineering & 5 (36\%) & 8 (57\%) & 1 (7\%) & 14 \\
    \bottomrule
  \end{tabular}
\end{table}

Petrophysics has the lowest share of easy questions (18\%) and the highest share of hard questions (24\%), consistent with the technical depth of well logging physics covered in Ellis \& Singer \cite{ellis_singer_2007}. Smaller domains such as Drilling Engineering and Production Engineering contain few hard questions, reflecting the limited coverage of these topics in the selected sources.

\FloatBarrier
\section{Evaluation setup}

\subsection{Models and providers}

The evaluation covers 72 models accessed through two API providers: Azure OpenAI \cite{azure_openai} and OpenRouter \cite{openrouter}. The evaluation ran in December 2025. Models come from OpenAI \cite{gpt4,gpt4o,gpt41,gpt5,o3_o4mini} (GPT-4o, GPT-4.1, GPT-5 series, o3-mini, o4-mini), Anthropic \cite{claude3,claude35,claude37,claude_opus45} (Claude 3.5 Haiku, Claude 3.7 Sonnet, Claude Opus 4.5, Claude Sonnet 4.5, Claude Haiku 4.5), Google \cite{gemini,gemini2,gemini25,gemini3,gemma3} (Gemini 2.0, 2.5, 3 series, Gemma 3), Meta \cite{llama3,llama31,llama32,llama4} (Llama 3.1, 3.2, 4), DeepSeek \cite{deepseek_v3,deepseek_v3_2,deepseek_r1} (R1, V3.2), Mistral \cite{mistral} (Small, Medium, Nemo, Ministral), Alibaba \cite{qwen3} (Qwen3 series), Zhipu \cite{glm4,glm47} (GLM-4, GLM-4.7), xAI \cite{grok3,grok} (Grok 3, 4), Moonshot \cite{kimi_k2} (Kimi K2), MiniMax \cite{minimax_m2} (M2), Microsoft \cite{phi4} (Phi-4) and Nvidia \cite{nemotron} (Nemotron).

Models range from compact 3B parameter variants to frontier reasoning models. Open-weight models (32 of 72) include GLM-4.7, DeepSeek-R1, Llama-4-Scout, Qwen3 variants, Mistral models and Gemma 3. Pricing spans from \$0.02/M input tokens (Llama-3.2-3b-instruct) to \$25/M output tokens (Claude Opus 4.5).


\subsection{Prompting and answer extraction}

The evaluation uses a zero-shot prompt format to assess model knowledge without providing examples.

The \textbf{system prompt} tells the model ``You are taking a multiple-choice exam on Oil \& Gas geoscience. For each question, select the single best answer from the options provided. State your final answer as a single letter: A, B, C or D.''

The \textbf{user prompt} presents the question text followed by four labeled choices (A--D) and ``Answer:'' as the final line.

\begin{figure}[htbp]
  \centering
  \includegraphics[width=\textwidth]{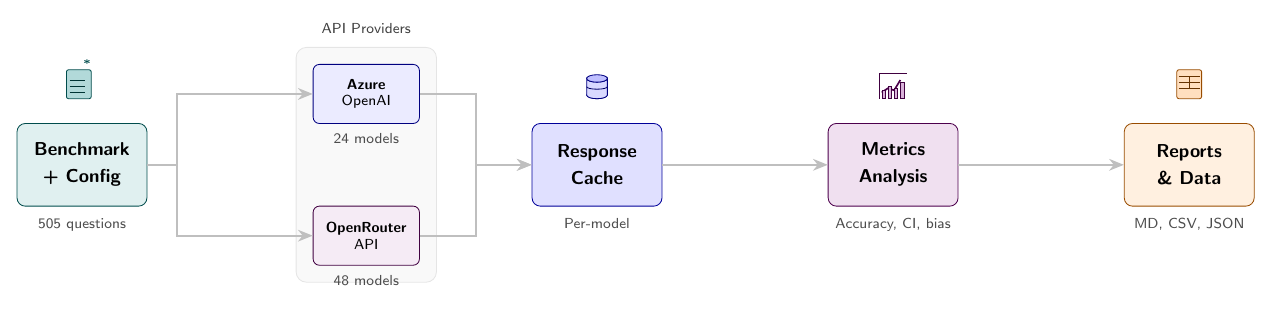}
  \caption{Evaluation pipeline. Models are configured via YAML, questions sent to
           Azure OpenAI or OpenRouter APIs, responses cached per model/question
           and analyzed to generate leaderboard and analysis reports.}
  \label{fig:eval_pipeline}
\end{figure}

Answer extraction uses flexible regex patterns to handle varied response formats. Preprocessing removes reasoning tags (\texttt{<think>}, \texttt{<thinking>}) from models like DeepSeek-R1 that expose chain-of-thought traces. The extraction logic prioritizes explicit patterns (``The answer is B'', ``Answer: C''). Failed extractions (where no A/B/C/D letter can be identified) are counted as incorrect answers. Figure~\ref{fig:eval_pipeline} shows the evaluation flow from configuration to report generation.


\subsection{Metrics}

The primary metric is overall accuracy, defined as correct answers divided by total questions (505).

Secondary metrics include accuracy by difficulty level (easy, medium, hard) and by domain (seven categories). The analysis also covers bias patterns: position bias (deviation from uniform A/B/C/D selection) and length bias (tendency to select the longest answer choice). The benchmark's residual length bias (correct answer is uniquely longest in 43.2\% of questions) provides a reference point for interpreting model length bias.


\subsection{Caching and reproducibility}

API responses are cached per model and question
({\small\texttt{eval/\allowbreak{}cache/\allowbreak{}\{model\}/\allowbreak{}\{question\_id\}.json}}). This enables resuming interrupted evaluation runs, re-analyzing results without additional API costs and debugging extraction failures.

Evaluation can be re-run in analyze-only mode to regenerate reports from cached responses. Output files include machine-readable JSON, Markdown tables and CSV exports with per-question breakdowns including raw model responses (truncated to 500 characters).


\section{Results}

\subsection{Overall results}

Table~\ref{tab:leaderboard} presents the top 50 models by accuracy \cite{formationeval_leaderboard}. The highest-performing model is Gemini 3 Pro Preview at 99.8\% (504/505 correct), followed by GLM-4.7 at 98.6\% and Gemini 3 Flash Preview at 98.2\%. Among open-weight models, GLM-4.7 leads, followed by DeepSeek-R1 (96.2\%) and DeepSeek-V3.2 (94.9\%).

\begin{figure}[htbp]
  \centering
  \includegraphics[width=\textwidth]{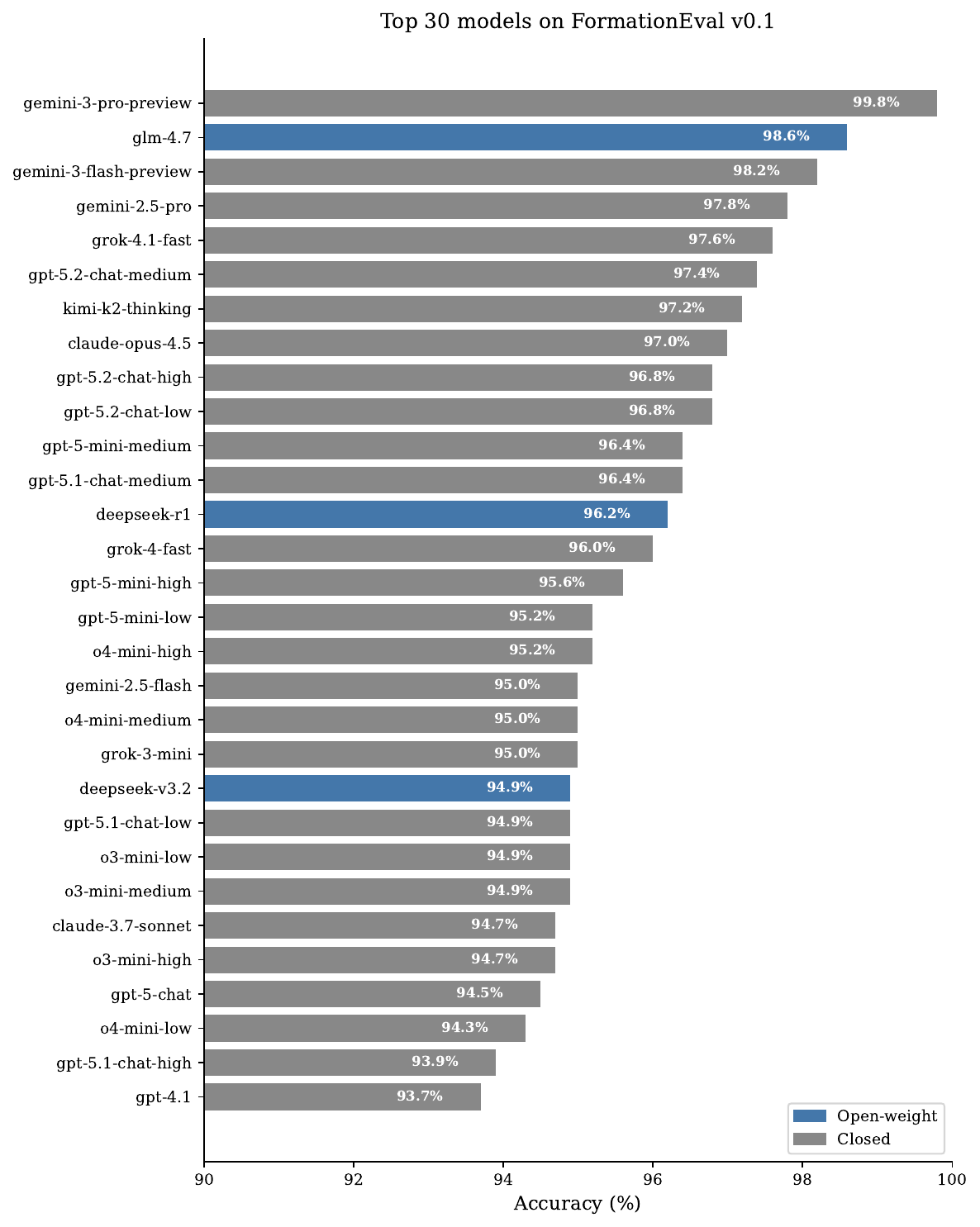}
  \caption{Top 30 models on FormationEval v0.1 by accuracy. Blue bars indicate open-weight models.
           GLM-4.7 (98.6\%) leads among open-weight models, ranking second overall.}
  \label{fig:top30_accuracy}
\end{figure}

\begin{figure}[htbp]
  \centering
  \includegraphics[width=\textwidth]{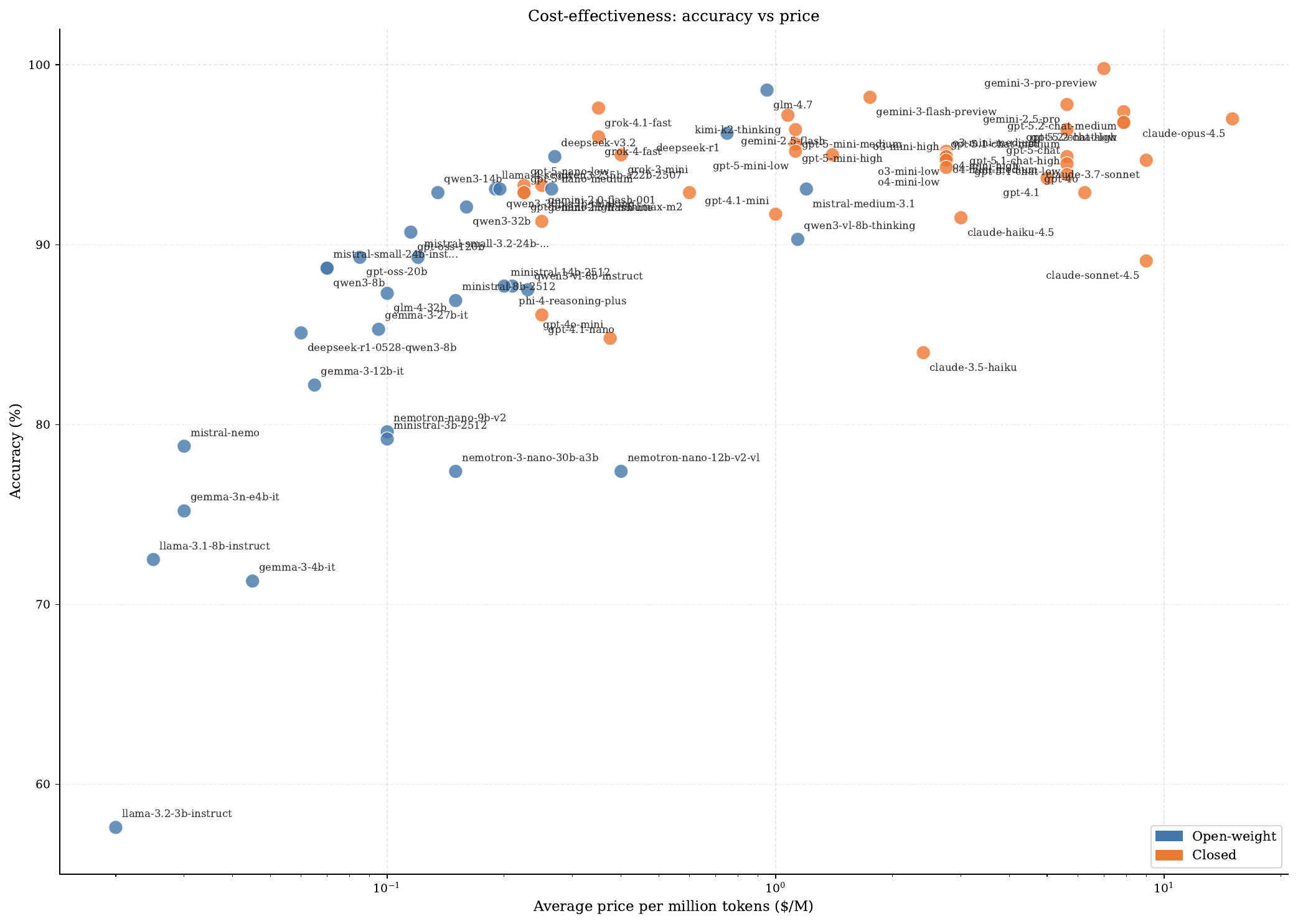}
  \caption{Cost-effectiveness analysis. Accuracy versus average token price (mean of input and output prices).
           Several high-accuracy models (Grok-4.1-fast, DeepSeek-R1) offer strong performance at lower cost.
           Open-weight models (blue) provide lower-cost alternatives to closed models (orange).}
  \label{fig:accuracy_vs_price}
\end{figure}

Accuracy spans a wide range, from 99.8\% (Gemini 3 Pro Preview) to 57.6\% (Llama-3.2-3b-instruct). Models from Google, OpenAI and Zhipu dominate the top positions (Figure~\ref{fig:top30_accuracy}). Pricing varies considerably, with some lower-cost models like Grok-4.1-fast (\$0.20/M input) achieving 97.6\% accuracy. Figure~\ref{fig:accuracy_vs_price} shows the cost-effectiveness trade-off across all models.


\begin{table}[htbp]
  \centering
  \caption{Top 50 models by accuracy (prices in USD per million tokens). See Appendix~\ref{app:full_leaderboard} for the complete 72-model leaderboard.}
  \label{tab:leaderboard}
  \small
  \begin{tabular}{rlllrr}
    \toprule
    Rank & Model & Open & \$/M (in/out) & Accuracy & Correct \\
    \midrule
    1 & gemini-3-pro-preview & No & 2.00/12.00 & 99.8\% & 504/505 \\
    2 & glm-4.7 & \textcolor{openweight}{Yes} & 0.40/1.50 & 98.6\% & 498/505 \\
    3 & gemini-3-flash-preview & No & 0.50/3.00 & 98.2\% & 496/505 \\
    4 & gemini-2.5-pro & No & 1.25/10.00 & 97.8\% & 494/505 \\
    5 & grok-4.1-fast & No & 0.20/0.50 & 97.6\% & 493/505 \\
    6 & gpt-5.2-chat-medium & No & 1.75/14.00 & 97.4\% & 492/505 \\
    7 & kimi-k2-thinking & No & 0.40/1.75 & 97.2\% & 491/505 \\
    8 & claude-opus-4.5 & No & 5.00/25.00 & 97.0\% & 490/505 \\
    9 & gpt-5.2-chat-high & No & 1.75/14.00 & 96.8\% & 489/505 \\
    10 & gpt-5.2-chat-low & No & 1.75/14.00 & 96.8\% & 489/505 \\
    11 & gpt-5-mini-medium & No & 0.25/2.00 & 96.4\% & 487/505 \\
    12 & gpt-5.1-chat-medium & No & 1.25/10.00 & 96.4\% & 487/505 \\
    13 & deepseek-r1 & \textcolor{openweight}{Yes} & 0.30/1.20 & 96.2\% & 486/505 \\
    14 & grok-4-fast & No & 0.20/0.50 & 96.0\% & 485/505 \\
    15 & gpt-5-mini-high & No & 0.25/2.00 & 95.6\% & 483/505 \\
    16 & gpt-5-mini-low & No & 0.25/2.00 & 95.2\% & 481/505 \\
    17 & o4-mini-high & No & 1.10/4.40 & 95.2\% & 481/505 \\
    18 & gemini-2.5-flash & No & 0.30/2.50 & 95.0\% & 480/505 \\
    19 & o4-mini-medium & No & 1.10/4.40 & 95.0\% & 480/505 \\
    20 & grok-3-mini & No & 0.30/0.50 & 95.0\% & 480/505 \\
    21 & deepseek-v3.2 & \textcolor{openweight}{Yes} & 0.22/0.32 & 94.9\% & 479/505 \\
    22 & gpt-5.1-chat-low & No & 1.25/10.00 & 94.9\% & 479/505 \\
    23 & o3-mini-low & No & 1.10/4.40 & 94.9\% & 479/505 \\
    24 & o3-mini-medium & No & 1.10/4.40 & 94.9\% & 479/505 \\
    25 & claude-3.7-sonnet & No & 3.00/15.00 & 94.7\% & 478/505 \\
    26 & o3-mini-high & No & 1.10/4.40 & 94.7\% & 478/505 \\
    27 & gpt-5-chat & No & 1.25/10.00 & 94.5\% & 477/505 \\
    28 & o4-mini-low & No & 1.10/4.40 & 94.3\% & 476/505 \\
    29 & gpt-5.1-chat-high & No & 1.25/10.00 & 93.9\% & 474/505 \\
    30 & gpt-4.1 & No & 2.00/8.00 & 93.7\% & 473/505 \\
    31 & gemini-2.0-flash-001 & No & 0.10/0.40 & 93.3\% & 471/505 \\
    32 & gpt-5-nano-low & No & 0.05/0.40 & 93.3\% & 471/505 \\
    33 & llama-4-scout & \textcolor{openweight}{Yes} & 0.08/0.30 & 93.1\% & 470/505 \\
    34 & mistral-medium-3.1 & \textcolor{openweight}{Yes} & 0.40/2.00 & 93.1\% & 470/505 \\
    35 & qwen3-235b-a22b-2507 & \textcolor{openweight}{Yes} & 0.07/0.46 & 93.1\% & 470/505 \\
    36 & qwen3-30b-a3b-thinking-2507 & \textcolor{openweight}{Yes} & 0.05/0.34 & 93.1\% & 470/505 \\
    37 & gpt-4o & No & 2.50/10.00 & 92.9\% & 469/505 \\
    38 & gpt-5-nano-high & No & 0.05/0.40 & 92.9\% & 469/505 \\
    39 & gpt-5-nano-medium & No & 0.05/0.40 & 92.9\% & 469/505 \\
    40 & minimax-m2 & No & 0.20/1.00 & 92.9\% & 469/505 \\
    41 & qwen3-14b & \textcolor{openweight}{Yes} & 0.05/0.22 & 92.9\% & 469/505 \\
    42 & qwen3-32b & \textcolor{openweight}{Yes} & 0.08/0.24 & 92.1\% & 465/505 \\
    43 & gpt-4.1-mini & No & 0.40/1.60 & 91.7\% & 463/505 \\
    44 & claude-haiku-4.5 & No & 1.00/5.00 & 91.5\% & 462/505 \\
    45 & gemini-2.5-flash-lite & No & 0.10/0.40 & 91.3\% & 461/505 \\
    46 & gpt-oss-120b & \textcolor{openweight}{Yes} & 0.04/0.19 & 90.7\% & 458/505 \\
    47 & qwen3-vl-8b-thinking & \textcolor{openweight}{Yes} & 0.18/2.10 & 90.3\% & 456/505 \\
    48 & mistral-small-3.2-24b-instruct & \textcolor{openweight}{Yes} & 0.06/0.18 & 89.3\% & 451/505 \\
    49 & gpt-oss-20b & \textcolor{openweight}{Yes} & 0.03/0.14 & 89.3\% & 451/505 \\
    50 & claude-sonnet-4.5 & No & 3.00/15.00 & 89.1\% & 450/505 \\
    \bottomrule
  \end{tabular}
\end{table}

\FloatBarrier
\subsection{By difficulty}

\begin{figure}[htbp]
  \centering
  \includegraphics[width=0.85\textwidth]{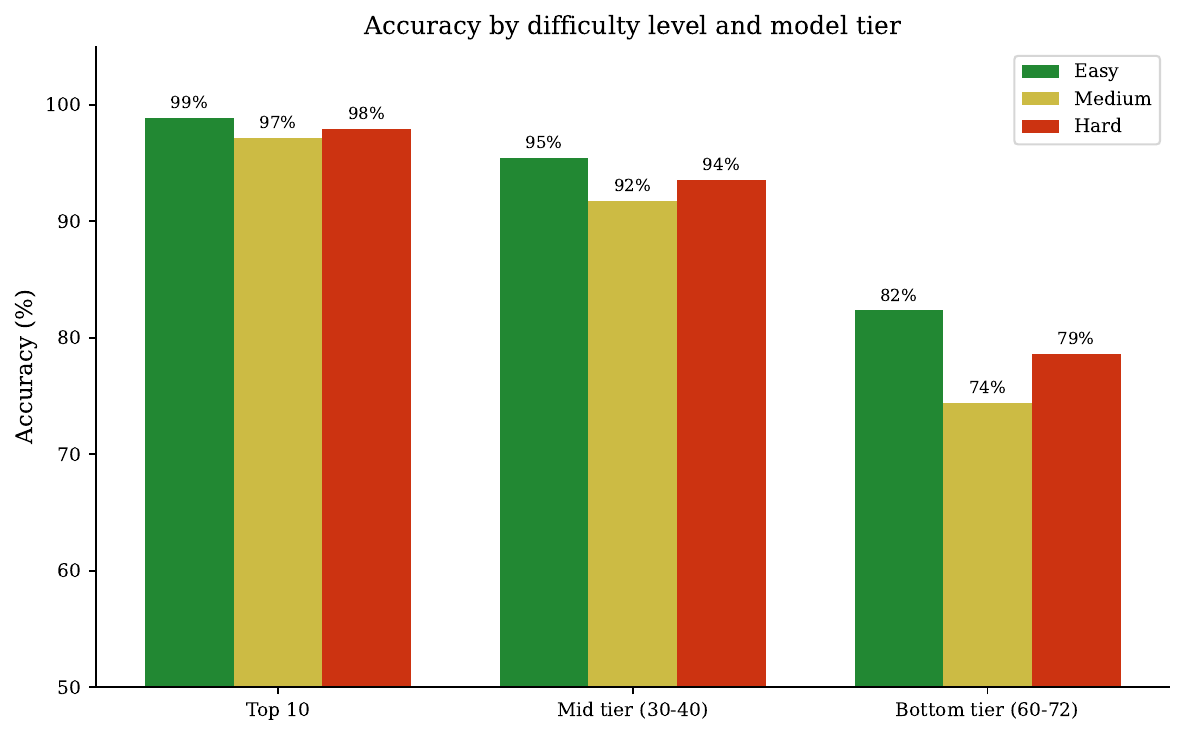}
  \caption{Accuracy by difficulty level across model tiers. Top-tier models show consistent performance across difficulty levels, while bottom-tier models exhibit larger gaps between easy and harder questions.}
  \label{fig:difficulty_breakdown}
\end{figure}

Across all model tiers, medium questions show the lowest accuracy, below both easy and hard (Figure~\ref{fig:difficulty_breakdown}). Even Gemini 3 Pro Preview made its single error on a medium question. Medium questions are more comparison-heavy and tool-specific, have a higher calculation rate (8.8\% versus 5.1\% for hard) and have less length-bias cueing than hard (correct answer is uniquely longest in 40.5\% of medium versus 55.6\% of hard, per the benchmark dataset \cite{formationeval_dataset}).

Smaller models show wider variance by difficulty. Llama-3.2-3b-instruct achieves 62.9\% on easy questions but only 54.7\% on medium, indicating difficulty labels correlate with model performance patterns. Figure~\ref{fig:difficulty_breakdown} compares accuracy by difficulty across model tiers.

\FloatBarrier
\subsection{By domain}

Petrophysics emerges as the most challenging domain across all models, with typical accuracy 3--5 percentage points lower than other domains. This reflects the technical detail of well logging physics and formation evaluation concepts.

The top model (Gemini 3 Pro Preview) achieves near-perfect scores across all domains: 99.6\% Petrophysics, 100\% for all other domains. Open-weight leader GLM-4.7 shows 98.2\% Petrophysics versus 99--100\% for other domains.

\begin{figure}[htbp]
  \centering
  \includegraphics[width=\textwidth]{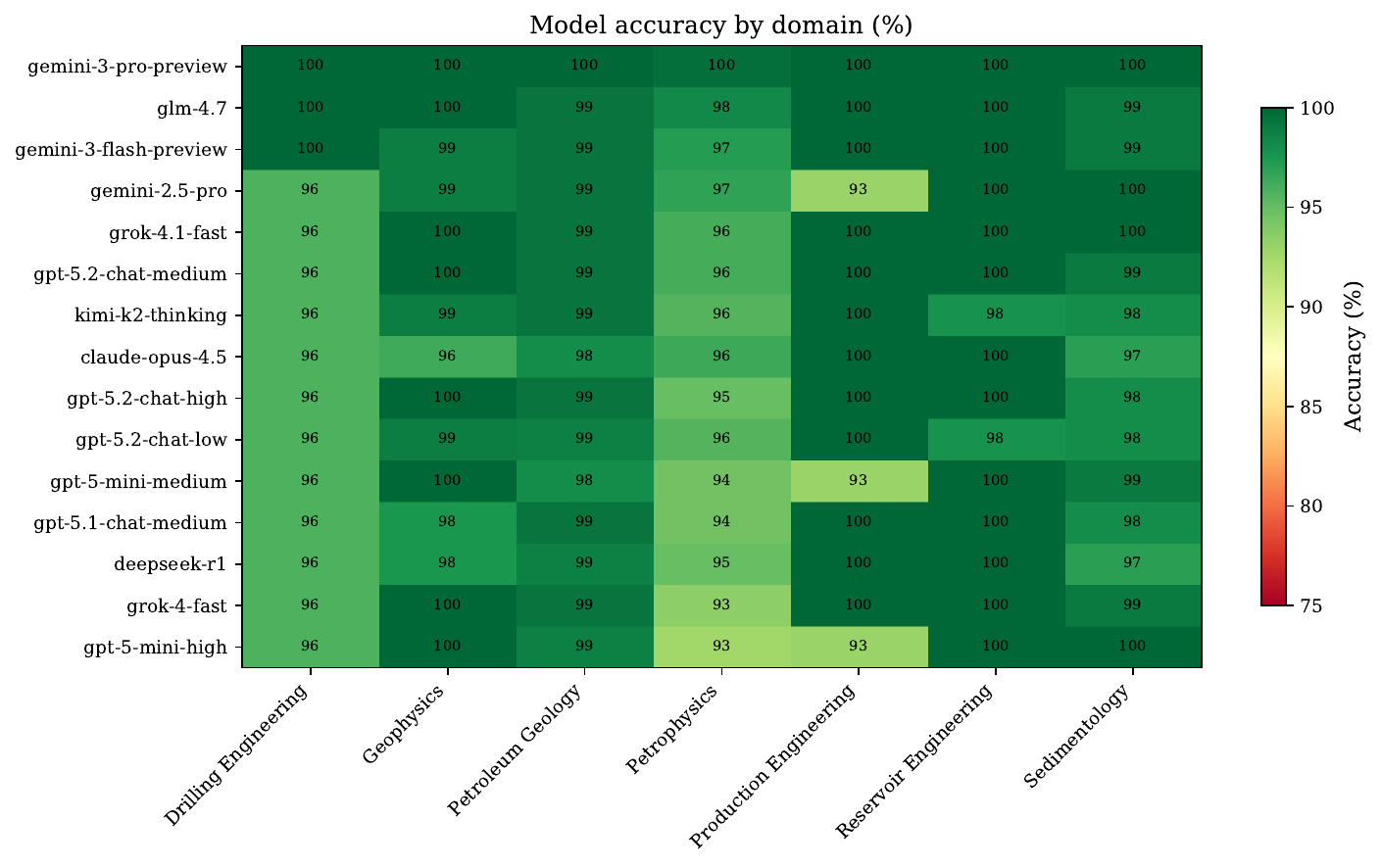}
  \caption{Model accuracy by domain for top 15 models. Petrophysics shows consistently lower accuracy than other domains across all models, reflecting the technical detail of well logging concepts.}
  \label{fig:domain_heatmap}
\end{figure}

Table~\ref{tab:domain_avg} shows average accuracy per domain across all 72 models. Reservoir Engineering leads at 95.6\%, while Petrophysics is consistently lowest at 87.5\%. Top models often achieve 100\% on Production and Drilling (which have fewer questions: 14 and 24 respectively), but overall averages place these domains in the middle of the ranking. Figure~\ref{fig:domain_heatmap} shows accuracy patterns for the top 15 models.

\begin{table}[htbp]
  \centering
  \caption{Average accuracy by domain across all 72 models.}
  \label{tab:domain_avg}
  \begin{tabular}{lr}
    \toprule
    Domain & Average accuracy \\
    \midrule
    Reservoir Engineering & 95.6\% \\
    Petroleum Geology & 93.9\% \\
    Sedimentology & 93.6\% \\
    Geophysics & 93.2\% \\
    Production Engineering & 91.5\% \\
    Drilling Engineering & 91.3\% \\
    Petrophysics & 87.5\% \\
    \bottomrule
  \end{tabular}
\end{table}

\FloatBarrier
\subsection{Hardest questions}

The ten hardest questions (by model failure rate) reveal systematic knowledge gaps \cite{formationeval_analysis}. The hardest question (on strike-slip fault stepovers and pull-apart basin formation) was answered incorrectly by 61 of 72 models (85\%). Most models selected option A (left-stepping arrangement) instead of the correct answer D (right-stepping arrangement for dextral motion). The top three hardest questions are documented in the analysis report \cite{formationeval_analysis}.

Eight of the ten hardest questions are from the Petrophysics domain, covering specialized topics like neutron-density crossplot interpretation, invasion profiles and tool calibration. One geology question (strike-slip stepovers) and one easy-labeled question (logging-while-drilling (LWD) propagation) also appear in the top 10.

Model agreement analysis shows that 21.6\% of questions were answered correctly by all 72 models, 0\% were missed by all models and 78.4\% showed mixed results, indicating most questions help distinguish model performance levels.


\FloatBarrier
\subsection{Open-weight models}

A key question for this evaluation was whether open-weight models, especially smaller and cheaper ones, have enough domain knowledge for petroleum geoscience. While larger closed models were expected to perform well, the capabilities of open-weight alternatives in this specialized field were less clear.

\begin{figure}[htbp]
  \centering
  \includegraphics[width=\textwidth]{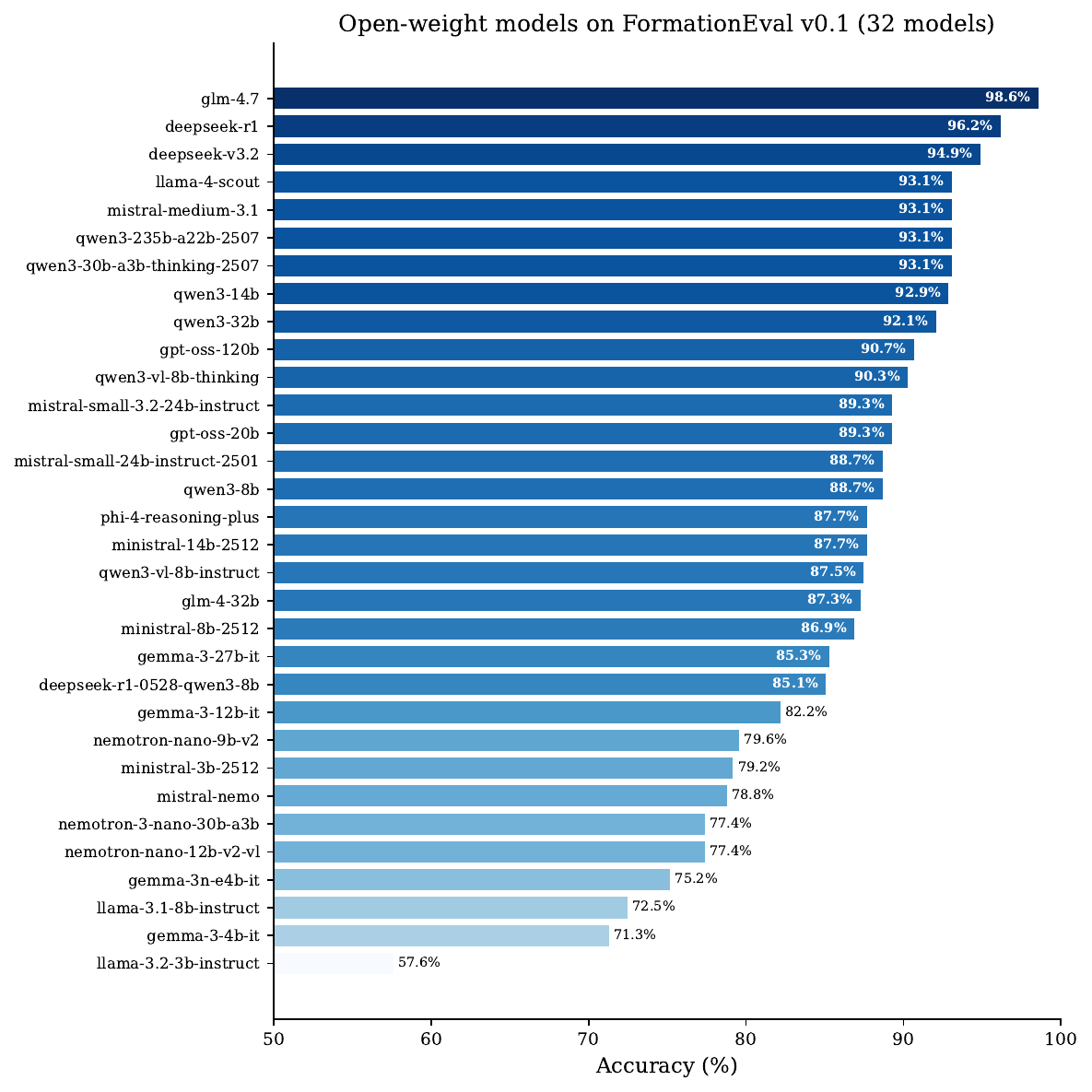}
  \caption{All 32 open-weight models ranked by accuracy. GLM-4.7 leads at 98.6\%, followed by DeepSeek-R1 (96.2\%) and DeepSeek-V3.2 (94.9\%). Color intensity indicates accuracy level.}
  \label{fig:open_weight_models}
\end{figure}

Of the 72 models evaluated, 32 have publicly available weights. Open-weight accuracy ranges from 57.6\% (Llama-3.2-3b-instruct) to 98.6\% (GLM-4.7), with mean 85.7\% and median 87.7\%. Eleven open-weight models reach at least 90\% and 22 reach at least 85\%. Figure~\ref{fig:open_weight_models} visualizes all open-weight models by accuracy; Table~\ref{tab:open_weight} provides the detailed breakdown.

The highest-accuracy open-weight models are GLM-4.7, DeepSeek-R1, DeepSeek-V3.2, Llama-4-Scout and Mistral Medium 3.1, followed by large Qwen3 variants at 93.1\%. Qwen3 base variants span 88.7--93.1\%, Qwen3-VL variants 87.5--90.3\% and GPT-OSS 89.3--90.7\%. Mistral and Ministral models span 78.8--93.1\%; GLM-4-32B reaches 87.3\%; Phi-4 reasoning-plus reaches 87.7\%; DeepSeek-R1-0528-Qwen3-8B reaches 85.1\%. DeepSeek-R1 at \$0.30/M input offers similar accuracy to GPT-5.1 variants at \$1.25/M input; Qwen3-14b reaches 92.9\% at \$0.05/M input.

Across domains, open-weight averages are highest in Reservoir Engineering (93.3\%) and lowest in Petrophysics (82.0\%), with other domains between 86.4\% and 90.2\% (Drilling 87.3\%, Geophysics 89.8\%, Petroleum Geology 90.2\%, Production 86.4\%, Sedimentology 89.6\%). GLM-4.7 achieves the top open-weight score in every domain, leading alone in four (Geophysics, Petroleum Geology, Petrophysics, Sedimentology) and tying with other models in three. The lowest scores across all domains come from Llama-3.2-3b-instruct, including 45.8\% in Drilling and 51.1\% in Petrophysics.

These results suggest that petroleum geoscience knowledge is well-represented in open-weight models, with 11 of 32 exceeding 90\% accuracy and 22 exceeding 85\%. Even models with fewer parameters and lower costs show useful domain understanding, making them viable options for applications where open weights or cost efficiency matter. Smaller closed models also perform well (GPT-5-nano variants achieve 92.9--93.3\%), suggesting that compact models of either type can handle this domain.

\begin{longtable}{rllrr}
  \caption{Open-weight models by accuracy (prices in USD per million tokens).} \label{tab:open_weight} \\
  \toprule
  Overall rank & Model & \$/M (in/out) & Accuracy & Correct \\
  \midrule
  \endfirsthead
  \multicolumn{5}{c}{\tablename\ \thetable{} -- continued from previous page} \\
  \toprule
  Overall rank & Model & \$/M (in/out) & Accuracy & Correct \\
  \midrule
  \endhead
  \midrule \multicolumn{5}{r}{Continued on next page} \\
  \endfoot
  \bottomrule
  \endlastfoot
  2 & glm-4.7 & 0.40/1.50 & 98.6\% & 498/505 \\
  13 & deepseek-r1 & 0.30/1.20 & 96.2\% & 486/505 \\
  21 & deepseek-v3.2 & 0.22/0.32 & 94.9\% & 479/505 \\
  33 & llama-4-scout & 0.08/0.30 & 93.1\% & 470/505 \\
  34 & mistral-medium-3.1 & 0.40/2.00 & 93.1\% & 470/505 \\
  35 & qwen3-235b-a22b-2507 & 0.07/0.46 & 93.1\% & 470/505 \\
  36 & qwen3-30b-a3b-thinking-2507 & 0.05/0.34 & 93.1\% & 470/505 \\
  41 & qwen3-14b & 0.05/0.22 & 92.9\% & 469/505 \\
  42 & qwen3-32b & 0.08/0.24 & 92.1\% & 465/505 \\
  46 & gpt-oss-120b & 0.04/0.19 & 90.7\% & 458/505 \\
  47 & qwen3-vl-8b-thinking & 0.18/2.10 & 90.3\% & 456/505 \\
  48 & mistral-small-3.2-24b-instruct & 0.06/0.18 & 89.3\% & 451/505 \\
  49 & gpt-oss-20b & 0.03/0.14 & 89.3\% & 451/505 \\
  51 & mistral-small-24b-instruct-2501 & 0.03/0.11 & 88.7\% & 448/505 \\
  52 & qwen3-8b & 0.03/0.11 & 88.7\% & 448/505 \\
  53 & phi-4-reasoning-plus & 0.07/0.35 & 87.7\% & 443/505 \\
  54 & ministral-14b-2512 & 0.20/0.20 & 87.7\% & 443/505 \\
  55 & qwen3-vl-8b-instruct & 0.06/0.40 & 87.5\% & 442/505 \\
  56 & glm-4-32b & 0.10/0.10 & 87.3\% & 441/505 \\
  57 & ministral-8b-2512 & 0.15/0.15 & 86.9\% & 439/505 \\
  59 & gemma-3-27b-it & 0.04/0.15 & 85.3\% & 431/505 \\
  60 & deepseek-r1-0528-qwen3-8b & 0.02/0.10 & 85.1\% & 430/505 \\
  63 & gemma-3-12b-it & 0.03/0.10 & 82.2\% & 415/505 \\
  64 & nemotron-nano-9b-v2 & 0.04/0.16 & 79.6\% & 402/505 \\
  65 & ministral-3b-2512 & 0.10/0.10 & 79.2\% & 400/505 \\
  66 & mistral-nemo & 0.02/0.04 & 78.8\% & 398/505 \\
  67 & nemotron-3-nano-30b-a3b & 0.06/0.24 & 77.4\% & 391/505 \\
  68 & nemotron-nano-12b-v2-vl & 0.20/0.60 & 77.4\% & 391/505 \\
  69 & gemma-3n-e4b-it & 0.02/0.04 & 75.2\% & 380/505 \\
  70 & llama-3.1-8b-instruct & 0.02/0.03 & 72.5\% & 366/505 \\
  71 & gemma-3-4b-it & 0.02/0.07 & 71.3\% & 360/505 \\
  72 & llama-3.2-3b-instruct & 0.02/0.02 & 57.6\% & 291/505 \\
\end{longtable}

\FloatBarrier
\subsection{Bottom performers}

The ten lowest-scoring models are all compact variants (3B--12B parameters) or older architectures. Llama-3.2-3b-instruct achieves 57.6\%, only marginally above random guessing (25\%). The bottom tier includes Gemma-3-4b-it (71.3\%), Llama-3.1-8b-instruct (72.5\%), Gemma-3n-e4b-it (75.2\%) and Nemotron variants (77--80\%).

These models show consistent patterns. Accuracy on easy questions (63--87\%) exceeds medium (55--78\%) and hard (59--82\%), but the gap between easy and medium is larger than for top models. Llama-3.2-3b-instruct shows the widest spread, at 62.9\% easy versus 54.7\% medium.

Petrophysics accuracy for bottom-tier models falls to 51--74\%, while other domains remain at 62--98\%. This suggests that smaller models particularly struggle with the technical detail of well logging concepts. The full 72-model leaderboard is provided in Appendix~\ref{app:full_leaderboard}.


\section{Discussion}

\subsection{Interpretation of scores}

The benchmark provides relative comparisons between models rather than absolute measures of domain competence. High scores indicate that a model can answer concept-based questions derived from authoritative sources, but do not guarantee expertise in practical applications.

All models exhibit elevated length bias in this analysis. They select the longest answer choice more often than the 25\% baseline. This rate (38--47\% across models) is close to the benchmark's residual bias (correct answer is uniquely longest in 43.2\% of questions), making it difficult to distinguish length exploitation from genuine reasoning that produces longer correct answers.

Position bias is generally low across models, with most showing near-uniform A/B/C/D distributions. Two Nvidia Nemotron variants showed elevated position bias (``High'' level), potentially indicating instruction-following limitations.

\subsection{Limitations and threats to validity}

\textbf{Residual length bias} persists despite mitigation efforts. Correct answers remain uniquely longest in 43.2\% of questions (versus 25\% expected), which may inflate scores for models sensitive to answer length.

\textbf{Contamination risk} cannot be fully ruled out. While questions are generated from concepts rather than copied, the underlying topics appear in textbooks that may be in model training data. Contamination risk labels are provided but actual training data overlap cannot be verified.

\textbf{Quality assurance} combined batch review of all 505 questions with spot checks against source material (Section~3.5), but did not include independent expert review. Despite 67 corrections identified during the process, additional errors may remain.

\textbf{Domain coverage} is uneven, with petrophysics dominating (54\% of questions) due to source availability. This may not reflect the breadth of petroleum geoscience equally.

\subsection{Provider and pricing variability}

Model pricing fluctuates and may not reflect values at time of reading. OpenRouter and Azure pricing differ for the same models. Some models are available through multiple providers at different costs.

Provider reliability varies. Rate limits, latency and availability differ across Azure OpenAI and OpenRouter endpoints. The caching strategy mitigates this for reproducibility but may not reflect real-world API behavior.


\section{Release and reproducibility}

The benchmark artifacts are organized for reproducibility. Version-controlled outputs include the dataset JSON and PDF, leaderboard and analysis reports in Markdown, and per-question CSV with raw model responses (truncated to 500 characters). Cached responses (gitignored) store raw API responses per model and question, enabling re-analysis without additional API calls. A PDF export of the full dataset provides question cards, metadata and bookmarks for domain expert review.

Evaluation scripts support an analyze-only mode that regenerates reports from cached responses without making API calls, ensuring reproducibility even when model APIs change or become unavailable.


\section{Data and code availability}
The benchmark, evaluation reports and code are available in the project repository \cite{formationeval_repo}.

\begin{itemize}
  \item Repository: \href{https://github.com/AlmazErmilov/FormationEval-an-Open-Benchmark-for-Oil-Gas-Geoscience-MCQ-Evaluation}{\texttt{github.com/AlmazErmilov/FormationEval-an-Open-Benchmark...}}
  \item Dataset JSON \cite{formationeval_dataset}: \href{https://github.com/AlmazErmilov/FormationEval-an-Open-Benchmark-for-Oil-Gas-Geoscience-MCQ-Evaluation/blob/v0.1/data/benchmark/formationeval_v0.1.json}{\texttt{data/benchmark/formationeval\_v0.1.json}}
  \item Dataset PDF \cite{formationeval_pdf}: \href{https://github.com/AlmazErmilov/FormationEval-an-Open-Benchmark-for-Oil-Gas-Geoscience-MCQ-Evaluation/blob/v0.1/data/benchmark/formationeval_v0.1.pdf}{\texttt{data/benchmark/formationeval\_v0.1.pdf}}
  \item Leaderboard: \href{https://github.com/AlmazErmilov/FormationEval-an-Open-Benchmark-for-Oil-Gas-Geoscience-MCQ-Evaluation/blob/v0.1/eval/results/leaderboard.md}{\texttt{eval/results/leaderboard.md}}
  \item Analysis: \href{https://github.com/AlmazErmilov/FormationEval-an-Open-Benchmark-for-Oil-Gas-Geoscience-MCQ-Evaluation/blob/v0.1/eval/results/analysis.md}{\texttt{eval/results/analysis.md}}
  \item Per-question results: \href{https://github.com/AlmazErmilov/FormationEval-an-Open-Benchmark-for-Oil-Gas-Geoscience-MCQ-Evaluation/blob/v0.1/eval/results/questions.csv}{\texttt{eval/results/questions.csv}}
  \item Generation scripts: \href{https://github.com/AlmazErmilov/FormationEval-an-Open-Benchmark-for-Oil-Gas-Geoscience-MCQ-Evaluation/tree/v0.1/src}{\texttt{src/}}
  \item Evaluation scripts: \href{https://github.com/AlmazErmilov/FormationEval-an-Open-Benchmark-for-Oil-Gas-Geoscience-MCQ-Evaluation/tree/v0.1/eval}{\texttt{eval/}}
  \item Interactive website and quiz \cite{formationeval_website}: \href{https://www.formationeval.no}{\texttt{formationeval.no}}
\end{itemize}

\section{Conclusion and future work}

This paper introduced FormationEval, a 505-question multiple-choice benchmark for evaluating language models on petroleum geoscience. The benchmark covers seven domains derived from authoritative sources using a concept-based methodology that respects copyright while testing domain knowledge.

Evaluation of 72 models reveals that frontier models achieve over 97\% accuracy, with Gemini 3 Pro Preview leading at 99.8\%. Open-weight models perform competitively, with GLM-4.7 achieving 98.6\%. Petrophysics emerges as the most challenging domain across all models.

Future work includes expanding to additional languages (Norwegian, Russian), adding questions from more sources to balance domain coverage and developing contamination detection methods. The benchmark is intended as a living resource, with new models added to the leaderboard as they become available.

\FloatBarrier
\clearpage
\appendix

\section{Schema reference}

Each question includes the following fields:

\vspace{1em}
\begin{center}
\small
\begin{tabular}{ll}
  \toprule
  Field & Description \\
  \midrule
  \texttt{id} & Unique identifier \\
  \texttt{version} & Version string (e.g., ``0.1'') \\
  \texttt{question} & Question text \\
  \texttt{choices} & Array of 4 options (A--D) \\
  \texttt{answer\_index} & Correct answer index (0--3) \\
  \texttt{answer\_key} & Correct answer letter (A--D) \\
  \texttt{rationale} & Explanation of correct answer \\
  \texttt{difficulty} & easy $|$ medium $|$ hard \\
  \texttt{language} & Language code: en $|$ ru $|$ no \\
  \texttt{domains} & Array of broad categories \\
  \texttt{topics} & Array of specific subjects \\
  \texttt{sources} & Provenance metadata array \\
  \texttt{derivation\_mode} & Always ``concept\_based'' \\
  \texttt{metadata.calc\_required} & Boolean (calculation needed) \\
  \texttt{metadata.contamination\_risk} & low $|$ medium $|$ high \\
  \bottomrule
\end{tabular}
\end{center}
\vspace{1em}

Contamination risk indicates likelihood that similar questions exist in LLM training data: \textit{low} for questions unique to this source, \textit{medium} for common concepts and \textit{high} for standard textbook topics.

\section{Prompt templates}

\subsection{Evaluation prompt}

\textbf{System prompt:}
\begin{quote}
\small\texttt{You are taking a multiple-choice exam on Oil \& Gas geoscience.}\\
\small\texttt{For each question, select the single best answer from the options provided.}\\
\small\texttt{State your final answer as a single letter: A, B, C or D.}
\end{quote}

\textbf{User prompt format:}
\begin{quote}
\texttt{\{question\}}\\[0.5em]
\texttt{A) \{choice\_a\}}\\
\texttt{B) \{choice\_b\}}\\
\texttt{C) \{choice\_c\}}\\
\texttt{D) \{choice\_d\}}\\[0.5em]
\texttt{Answer:}
\end{quote}

\subsection{Multiple-choice question generation prompt summary}

The full system prompt for MCQ generation (475 lines) is \href{https://github.com/AlmazErmilov/FormationEval-an-Open-Benchmark-for-Oil-Gas-Geoscience-MCQ-Evaluation/blob/v0.1/src/prompts/mcq_generator_system_prompt.txt}{available in the repository}.
Key instructions include:

\begin{enumerate}
\item Generate questions that test \textbf{understanding of concepts}, not recognition of phrases
\item Test one concept per question
\item Never copy sentences or descriptive phrases from source text
\item Create standalone questions answerable from domain knowledge without access to the source chapter
\item Cover key concepts without repetition
\item Distribute difficulty: 30\% easy, 50\% medium, 20\% hard
\item Balance answer options in length and structure
\item Distribute correct answers evenly across positions A/B/C/D
\item Avoid ``All of the above'' and ``None of the above''
\item Avoid negative phrasing (``Which is NOT...'')
\item Avoid exploitable patterns with qualifier words
\item Provide rationale explaining why the answer is correct
\item Include source metadata and contamination risk assessment
\end{enumerate}

\section{Bias mitigation summary}

Tables~\ref{tab:length_bias} and \ref{tab:qualifier_bias} summarize the bias mitigation efforts.

\begin{table}[htbp]
  \centering
  \caption{Length bias before and after mitigation.}
  \label{tab:length_bias}
  \begin{tabular}{lrr}
    \toprule
    Metric & Original & After fixes \\
    \midrule
    Correct answer is uniquely longest & $>$55\% & 43.2\% \\
    Correct answer avg length & 86.6 chars & 86.6 chars \\
    Distractor avg length & 69.8 chars & 74.0 chars \\
    \bottomrule
  \end{tabular}
\end{table}

\begin{table}[htbp]
  \centering
  \caption{Qualifier word patterns before and after mitigation.}
  \label{tab:qualifier_bias}
  \begin{tabular}{lrrrl}
    \toprule
    Word & In correct & In distractor & Correct rate & Status \\
    \midrule
    ``always'' & 0 & 49 & 0\% & Replaced with synonyms \\
    ``invariably'' & 0 & 12 & 0\% & New (from ``always'') \\
    ``necessarily'' & 0 & 14 & 0\% & New (from ``always'') \\
    ``inherently'' & 0 & 12 & 0\% & New (from ``always'') \\
    ``consistently'' & 0 & 9 & 0\% & New (from ``always'') \\
    ``may'' & 13 & 0 & 100\% & Original (pre-fix) \\
    ``may'' & 13 & 13 & 50\% & After fix \\
    \bottomrule
  \end{tabular}
\end{table}

\textbf{Mitigation applied}: (1) All 49 ``always'' instances replaced with varied synonyms to break single-word exploit; (2) Added ``may'' to 13 distractors with ``no-effect'' claims.

\textbf{Residual issues}: Absolute word synonyms still have 0\% correct rate. Combined ``any-absolute-word=wrong'' heuristic remains partially exploitable.

\section{Full model leaderboard}
\label{app:full_leaderboard}

Table~\ref{tab:full_leaderboard} presents all 72 evaluated models ranked by accuracy.

\begin{longtable}{rlllrr}
  \caption{Complete leaderboard. All 72 models by accuracy (prices in USD per million tokens).} \label{tab:full_leaderboard} \\
  \toprule
  Rank & Model & Open & \$/M (in/out) & Accuracy & Correct \\
  \midrule
  \endfirsthead
  \multicolumn{6}{c}{\tablename\ \thetable{} -- continued from previous page} \\
  \toprule
  Rank & Model & Open & \$/M (in/out) & Accuracy & Correct \\
  \midrule
  \endhead
  \midrule \multicolumn{6}{r}{Continued on next page} \\
  \endfoot
  \bottomrule
  \endlastfoot
  1 & gemini-3-pro-preview & No & 2.00/12.00 & 99.8\% & 504/505 \\
  2 & glm-4.7 & Yes & 0.40/1.50 & 98.6\% & 498/505 \\
  3 & gemini-3-flash-preview & No & 0.50/3.00 & 98.2\% & 496/505 \\
  4 & gemini-2.5-pro & No & 1.25/10.00 & 97.8\% & 494/505 \\
  5 & grok-4.1-fast & No & 0.20/0.50 & 97.6\% & 493/505 \\
  6 & gpt-5.2-chat-medium & No & 1.75/14.00 & 97.4\% & 492/505 \\
  7 & kimi-k2-thinking & No & 0.40/1.75 & 97.2\% & 491/505 \\
  8 & claude-opus-4.5 & No & 5.00/25.00 & 97.0\% & 490/505 \\
  9 & gpt-5.2-chat-high & No & 1.75/14.00 & 96.8\% & 489/505 \\
  10 & gpt-5.2-chat-low & No & 1.75/14.00 & 96.8\% & 489/505 \\
  11 & gpt-5-mini-medium & No & 0.25/2.00 & 96.4\% & 487/505 \\
  12 & gpt-5.1-chat-medium & No & 1.25/10.00 & 96.4\% & 487/505 \\
  13 & deepseek-r1 & Yes & 0.30/1.20 & 96.2\% & 486/505 \\
  14 & grok-4-fast & No & 0.20/0.50 & 96.0\% & 485/505 \\
  15 & gpt-5-mini-high & No & 0.25/2.00 & 95.6\% & 483/505 \\
  16 & gpt-5-mini-low & No & 0.25/2.00 & 95.2\% & 481/505 \\
  17 & o4-mini-high & No & 1.10/4.40 & 95.2\% & 481/505 \\
  18 & gemini-2.5-flash & No & 0.30/2.50 & 95.0\% & 480/505 \\
  19 & o4-mini-medium & No & 1.10/4.40 & 95.0\% & 480/505 \\
  20 & grok-3-mini & No & 0.30/0.50 & 95.0\% & 480/505 \\
  21 & deepseek-v3.2 & Yes & 0.22/0.32 & 94.9\% & 479/505 \\
  22 & gpt-5.1-chat-low & No & 1.25/10.00 & 94.9\% & 479/505 \\
  23 & o3-mini-low & No & 1.10/4.40 & 94.9\% & 479/505 \\
  24 & o3-mini-medium & No & 1.10/4.40 & 94.9\% & 479/505 \\
  25 & claude-3.7-sonnet & No & 3.00/15.00 & 94.7\% & 478/505 \\
  26 & o3-mini-high & No & 1.10/4.40 & 94.7\% & 478/505 \\
  27 & gpt-5-chat & No & 1.25/10.00 & 94.5\% & 477/505 \\
  28 & o4-mini-low & No & 1.10/4.40 & 94.3\% & 476/505 \\
  29 & gpt-5.1-chat-high & No & 1.25/10.00 & 93.9\% & 474/505 \\
  30 & gpt-4.1 & No & 2.00/8.00 & 93.7\% & 473/505 \\
  31 & gemini-2.0-flash-001 & No & 0.10/0.40 & 93.3\% & 471/505 \\
  32 & gpt-5-nano-low & No & 0.05/0.40 & 93.3\% & 471/505 \\
  33 & llama-4-scout & Yes & 0.08/0.30 & 93.1\% & 470/505 \\
  34 & mistral-medium-3.1 & Yes & 0.40/2.00 & 93.1\% & 470/505 \\
  35 & qwen3-235b-a22b-2507 & Yes & 0.07/0.46 & 93.1\% & 470/505 \\
  36 & qwen3-30b-a3b-thinking-2507 & Yes & 0.05/0.34 & 93.1\% & 470/505 \\
  37 & gpt-4o & No & 2.50/10.00 & 92.9\% & 469/505 \\
  38 & gpt-5-nano-high & No & 0.05/0.40 & 92.9\% & 469/505 \\
  39 & gpt-5-nano-medium & No & 0.05/0.40 & 92.9\% & 469/505 \\
  40 & minimax-m2 & No & 0.20/1.00 & 92.9\% & 469/505 \\
  41 & qwen3-14b & Yes & 0.05/0.22 & 92.9\% & 469/505 \\
  42 & qwen3-32b & Yes & 0.08/0.24 & 92.1\% & 465/505 \\
  43 & gpt-4.1-mini & No & 0.40/1.60 & 91.7\% & 463/505 \\
  44 & claude-haiku-4.5 & No & 1.00/5.00 & 91.5\% & 462/505 \\
  45 & gemini-2.5-flash-lite & No & 0.10/0.40 & 91.3\% & 461/505 \\
  46 & gpt-oss-120b & Yes & 0.04/0.19 & 90.7\% & 458/505 \\
  47 & qwen3-vl-8b-thinking & Yes & 0.18/2.10 & 90.3\% & 456/505 \\
  48 & mistral-small-3.2-24b-instruct & Yes & 0.06/0.18 & 89.3\% & 451/505 \\
  49 & gpt-oss-20b & Yes & 0.03/0.14 & 89.3\% & 451/505 \\
  50 & claude-sonnet-4.5 & No & 3.00/15.00 & 89.1\% & 450/505 \\
  51 & mistral-small-24b-instruct-2501 & Yes & 0.03/0.11 & 88.7\% & 448/505 \\
  52 & qwen3-8b & Yes & 0.03/0.11 & 88.7\% & 448/505 \\
  53 & phi-4-reasoning-plus & Yes & 0.07/0.35 & 87.7\% & 443/505 \\
  54 & ministral-14b-2512 & Yes & 0.20/0.20 & 87.7\% & 443/505 \\
  55 & qwen3-vl-8b-instruct & Yes & 0.06/0.40 & 87.5\% & 442/505 \\
  56 & glm-4-32b & Yes & 0.10/0.10 & 87.3\% & 441/505 \\
  57 & ministral-8b-2512 & Yes & 0.15/0.15 & 86.9\% & 439/505 \\
  58 & gpt-4.1-nano & No & 0.10/0.40 & 86.1\% & 435/505 \\
  59 & gemma-3-27b-it & Yes & 0.04/0.15 & 85.3\% & 431/505 \\
  60 & deepseek-r1-0528-qwen3-8b & Yes & 0.02/0.10 & 85.1\% & 430/505 \\
  61 & gpt-4o-mini & No & 0.15/0.60 & 84.8\% & 428/505 \\
  62 & claude-3.5-haiku & No & 0.80/4.00 & 84.0\% & 424/505 \\
  63 & gemma-3-12b-it & Yes & 0.03/0.10 & 82.2\% & 415/505 \\
  64 & nemotron-nano-9b-v2 & Yes & 0.04/0.16 & 79.6\% & 402/505 \\
  65 & ministral-3b-2512 & Yes & 0.10/0.10 & 79.2\% & 400/505 \\
  66 & mistral-nemo & Yes & 0.02/0.04 & 78.8\% & 398/505 \\
  67 & nemotron-3-nano-30b-a3b & Yes & 0.06/0.24 & 77.4\% & 391/505 \\
  68 & nemotron-nano-12b-v2-vl & Yes & 0.20/0.60 & 77.4\% & 391/505 \\
  69 & gemma-3n-e4b-it & Yes & 0.02/0.04 & 75.2\% & 380/505 \\
  70 & llama-3.1-8b-instruct & Yes & 0.02/0.03 & 72.5\% & 366/505 \\
  71 & gemma-3-4b-it & Yes & 0.02/0.07 & 71.3\% & 360/505 \\
  72 & llama-3.2-3b-instruct & Yes & 0.02/0.02 & 57.6\% & 291/505 \\
\end{longtable}

\section*{Declaration of Generative AI and AI-assisted technologies in the writing process}

During the preparation of this work, the author used GPT-5.2 (OpenAI) with extra high reasoning effort to transform textbook chapter content into multiple-choice questions following a structured generation pipeline (Section~3.4). All domain knowledge was derived from the source materials. The model served as a generation and reformatting tool, not as a knowledge source. GenAI tools were also used for language editing. The author reviewed and edited all generated content and takes full responsibility for the publication.

\bibliographystyle{plainurl}
\bibliography{refs}

\end{document}